\documentclass[runningheads]{llncs}
\usepackage[T1]{fontenc}
\usepackage{graphicx}
\usepackage{multirow}
\usepackage[table,xcdraw]{xcolor}
\begin{document}
\title{Test-time augmentation-based active learning and self-training for label-efficient segmentation}
\titlerunning{Test-time augmentation-based Active Learning and Self-training}
%
\author{Bella Specktor-Fadida\inst{1}
\and Anna Levchakov\inst{2} \and
Dana Schonberger\inst{2,3} \and Liat Ben-Sira \inst{4,5} 
\and Dafna Ben-Bashat \inst{2,5,6}
\and Leo Joskowicz \inst{1} 
}
\authorrunning{Specktor-Fadida et al.}
%
\institute{School of Computer Science and Eng., The Hebrew University of Jerusalem, Israel \and
Sagol Brain Institute, Tel Aviv Sourasky Medical Center, Israel \and Department of Biomedical Engineering, Tel Aviv University, Israel \and Division of Pediatric Radiology, Tel Aviv Sourasky Medical Center, Tel Aviv, Israel \and
Sackler Faculty of Medicine, Tel Aviv University, Israel \and Sagol School of Neuroscience, Tel Aviv University, Israel  }
\maketitle              
\begin{abstract}
Deep learning techniques depend on large datasets whose annotation is time-consuming. To reduce annotation burden, the self-training (ST) and active-learning (AL) methods have been developed as well as methods that combine them in an iterative fashion. However, it remains unclear when each method is the most useful, and when it is advantageous to combine them. In this paper, we propose a new method that combines ST with AL using Test-Time Augmentations (TTA). First, TTA is performed on an initial teacher network. Then, cases for annotation are selected based on the lowest estimated Dice score. Cases with high estimated scores are used as soft pseudo-labels for ST. The selected annotated cases are trained with existing annotated cases and ST cases with border slices annotations. We demonstrate the method on MRI fetal body and placenta segmentation tasks with different data variability characteristics. Our results indicate that ST is highly effective for both tasks, boosting performance for in-distribution (ID) and out-of-distribution (OOD) data. However, while self-training improved the performance of single-sequence fetal body segmentation when combined with AL, it slightly deteriorated performance of multi-sequence placenta segmentation on ID data. AL was helpful for the high variability placenta data, but did not improve upon random selection for the single-sequence body data. For fetal body segmentation sequence transfer, combining AL with ST following ST iteration yielded a Dice of 0.961 with only 6 original scans and 2 new sequence scans. Results using only 15 high-variability placenta cases were similar to those using 50 cases. Code is available at: \url{https://github.com/Bella31/TTA-quality-estimation-ST-AL}

\end{abstract}
\section{Introduction}
Segmentation of anatomical structures in CT and MRI scans is a key task in many clinical applications. However, manual annotation requires expertise, is tedious, time-consuming, and is subject to annotator variability. State-of-the-art automatic volumetric segmentation methods are based on deep neural networks. While effective, these methods require a large, high-quality dataset of expert-validated annotations, which is difficult to obtain. A variety of methods have been proposed to address annotation scarcity in semantic segmentation \cite{tajbakhsh2020embracing}.

Self-training (ST), also called pseudo-labeling, is a method for iterative semi-supervised learning (SSL). In ST, a model improves the quality of the pseudo-annotations by learning from its own predictions \cite{cheplygina2019not}. It is performed by alternatively training two or more networks with an uncertainty-aware scheme~\cite{cheplygina2019not,shen2023survey,tajbakhsh2020embracing}. Nie et al.~\cite{nie2018asdnet} describe an adversarial framework with a generator segmentation network and a discriminator confidence network. Xia et al.~\cite{xia2020uncertainty} propose a method for generating pseudo-labels based on co-training multiple networks. Consistency regularization (CR) \cite{chen2021semi,ouali2020semi}, another popular SSL method, relies on various perturbation techniques to generate disagreement on the same inputs, so that models can be trained by enforcing prediction consistency on unlabeled data without knowing the labeled information.

Active learning (AL) aims to identify the best samples for annotation during training. The most common criterion for sample selection is based on prediction uncertainty estimation, following the premise that more information can be learned by the model using uncertain samples \cite{budd2021survey}. AL often uses ensembled uncertainty measurements that require training multiple models or Monte-Carlo dropout~\cite{jungo2019assessing,yang2017suggestive}. Ensemble-based methods measure the agreement between different models, with higher model disagreement potentially implying more uncertainty and less informativeness. Jungo et al.~\cite{jungo2019assessing} indicated the need for subject-level uncertainty for quality estimation.

Segmentation quality estimation is used in ST to discard low-quality pseudo-labels and in AL to select low-quality samples for manual annotation. Wang et al.~\cite{wang2019aleatoric} show that aleatory uncertainty scoring with Test-Time Augmentations (TTA) provides a better uncertainty estimation than test-time dropout-based model uncertainty. They show that TTA helps to eliminate overconfident incorrect predictions. Specktor et al.~\cite{specktor2021bootstrap} discard low-quality pseudo-labels for ST with TTA-based Dice estimation. Others use TTA-based error estimation for AL with the Volume Overlap Distance and the Jensen-Shannon divergence~\cite{dudovitch2020deep,gaillochet2022taal}.

Frameworks that combine SSL and AL have been recently proposed for both natural and medical images~\cite{fathi2011combining,gaillochet2022taal,guan2023iterative,lai2021joint,nath2022warm,zhang2022boostmis}. Zhang et al. \cite{zhang2022boostmis} propose a closed loop pseudo-labeling and informative active annotation framework. Nath et al. \cite{nath2022warm} describe a two-stage learning framework for each active iteration where the unlabeled data is used in the second stage for semi-supervised fine-tuning. Gaillochet et al. \cite{gaillochet2022taal} applies cross-augmentation consistency during training and inference at each iteration to both improve model learning in a semi-supervised fashion and identify the most relevant unlabeled samples to annotate next. However, these methods do not take into account the data properties at each iteration. A key question remains whether it is always best to combine SSL and AL, and when it is better to use each separately.

Recently, an offline self-training scheme with one or two teacher-student iterations has been reported to yield superior segmentation results ~\cite{yang2022st++,zoph2020rethinking}. It consists of teacher training on annotated data followed by pseudo-labels generation on unlabeled data and student training that jointly learns from the manual and the high-quality generated pseudo-labels.

In this paper we present a new human-in-the-loop method based on TTA for a combined training of actively selected and offline-calculated self-trained examples. It performs simultaneous computation of pseudo-labels for self-training and segmentation quality estimation for an active selection of samples and filtering of unreliable pseudo-labels. The method also uses border slices annotations to improve the pseudo-labels and the quality estimation accuracy.

\section{Method}
The method consists of: 1) training a teacher network on a small number of examples; 2) training a student model using a fully-automatic ST module (optional); 3) training a second student model with a combined AL and ST module using the base network from step 1 or 2. The combined AL and ST iteration is performed either directly after the initial model training or after ST iteration.

\subsection{ST module}
The fully automatic ST iteration is optionally performed before the combined AL and ST module iteration. It uses soft teacher predictions instead of hard teacher predictions, as these were found to perform better~\cite{arazo2020pseudo,zhu2021improving}. See the Supplementary Material (Section 1, Table A1). The pseudo-labels are either network predictions without post-processing (soft network predictions) or TTA predictions medians for each voxel, depending on the benefit of using TTA for a given segmentation task. Unreliable training examples are filtered out with quality estimation ranking using a threshold. The threshold is selected so that the number of pseudo-labels will be at least the same as the number of labeled examples. 

The student self-training network is trained with the original data of the initial network and with the new self-training examples using an optimization scheme with learning rate restarts \cite{loshchilov2016sgdr}.

\subsubsection{TTA-based quality estimation:}The quality estimation is obtained by computing the mean Dice score between the median prediction and each of the augmentation results \cite{specktor2021bootstrap}. For TTA we use flipping, rotation, transpose and contrast augmentations. A median prediction is first assigned for each voxel, and then the estimated Dice score is obtained by computing the median Dice score between the median and each one of the augmentation results. 

\begin{figure}[t]
\centering
\includegraphics[height=6.5cm, width=12.1cm]{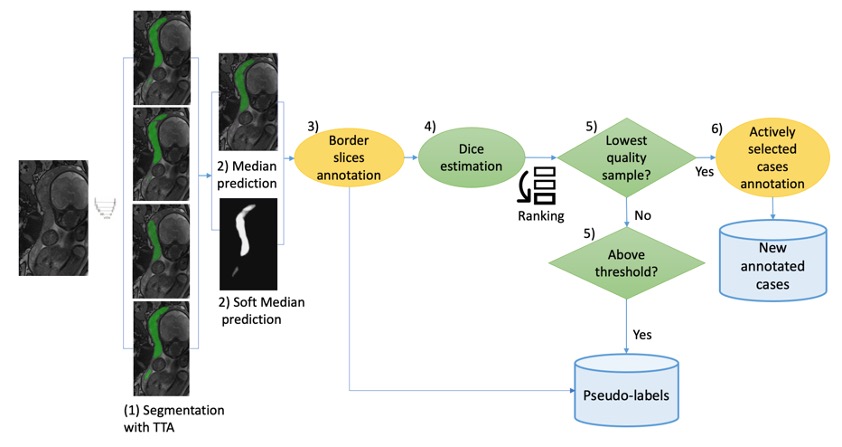}\\[-2ex]
\caption{Flow diagram of the combined active learning and self-training method illustrated on the placenta (dark green). The human annotation steps are marked in yellow; light green steps are automatic.}
\label{fig:ac_sl_method}
\end{figure}

\subsection{Combined AL and ST module}
The combined AL and ST module consists of five steps: 1) TTA inference is performed with a base network; 2) border slices of the region of Interest (ROI) are annotated for the TTA inference data; 3) cases are selected for annotation and unreliable pseudo-labels are filtered using TTA-based quality estimation on the slices of the structure of interest; 4) manual correction of the selected cases is performed for annotation; and 5) training is performed with the newly selected annotated cases, existing annotated cases, and additional ST cases.

The new network uses the original data in combination with new actively selected cases and pseudo-labels. Let $D$ be the set of all cases. The training set $D_{train}\subset D$ is $D_{train} = D_{base}\cup D_{AL}\cup D_{ST}$, where $D_{base}$ is the set of training examples of the initial network, $D_{AL}$ is the set of actively selected examples, and $D_{ST}$ is the set of selected pseudo-labeled examples. Training is performed by fine-tuning the base network with learning rate restarts optimization \cite{loshchilov2016sgdr}.

Fig.~\ref{fig:ac_sl_method} illustrates the process of generating the new training datasets $D_{AL}$ and $D_{ST}$. The process includes: 1) segmentation inference with TTA using the base network; 2) calculation of the median of TTA results for each voxel (soft median prediction) and thresholding with t=0.5 to produce the final median prediction mask (median prediction); 3) manual annotation of border slices; 4) TTA Dice estimation and ranking of new unlabeled cases by estimated Dice metric; 5) selection of cases with the lowest estimated Dice for annotation, and from the rest of the cases selection of pseudo labels above an estimated Dice threshold; 6) manual annotation of the actively selected cases.

Both the AL and ST cases use the quality estimation method described in [10], similar to the quality estimation in the fully automatic self-training module. The cases with the lowest estimated Dice score are selected for annotation and cases with estimated quality above a threshold are used as pseudo-labels. See the Supplementary Material for illustrations (Section 2, Fig. A1-A2).

\subsection{Border Slices Annotation} 
For this human-in-the-loop method, various types of manual annotations are performed on the unlabeled cases. To decrease errors in the background slices outside of the structure of interest, the uppermost and lowermost slices of the ROI that includes the structure of interest are manually selected by the annotator \cite{fadida2022partial}. This is an easy task that, and unlike segmentation annotations, can be quickly performed for a large number of cases. These annotations are used to correct the self-training result outside of the structure of interest and to estimate more accurately the Dice score in the selected region.

\section{Experimental Results}
To evaluate our method, we retrospectively collected fetal MRI scans and conducted three studies. The fetal MRI body scans were acquired with the FIESTA and the TRUFI sequences as part of routine fetal assessment of fetuses with gestational ages (GA) of 28-39 weeks from the Tel-Aviv Sourasky Medical Center, Israel. They included the segmentation of the fetal body or the fetal placenta. \\[2ex]
\noindent
{\bf Datasets and Annotations:}
The FIESTA sequence fetal body dataset consists of 104 cases acquired on a GE MR450 1.5T scanner with a resolution range of $1.48-1.87\times1.48-1.87\times2-5.0$ $mm^{3}$. The TRUFI sequence fetal body dataset consists of 101 cases acquired on Siemens Skyra 3T, Prisma 3T, and Aera 1.5T scanners, with a resolution range of $0.6-1.34\times0.6-1.34\times2-4.8$ $mm^{3}$. 

The placenta in-distribution (ID) dataset includes 88 cases, 50 TRUFI and 38 FIESTA. The TRUFI scans were acquired on Siemens Prisma and Vida scanners with a resolution of $0.781\times0.781\times2$ $mm^{3}$. The FIESTA sequence scans were acquired on a GE MR450 1.5T scanner with a resolution range of $1.48-1.87\times1.48-1.87\times2-5.0$ $mm^{3}$. For placenta segmentation, an additional Out-Of-Distribution (OOD) dataset with 24 TRUFI scans and restricted field of view not including the entire placenta was used. The OOD cases were acquired with Siemens Prisma and Vida scanners with a resolution of $0.586\times0.586\times4$ $mm^{3}$.

Both the annotations and the corrections were performed by a clinical trainee. All segmentations were validated by an expert fetal radiologist with >20 years of experience in fetal imaging.\\[2ex]
\noindent
{\bf Studies:}
We conducted three studies, two with fetal body segmentation data and one with placenta data. To address the high variability in segmentation quality of the low-data regime, we performed all the experiments with four different randomizations and averaged between them. In all experiments, we used 16 TTA variations. We evaluated the experiments with the Dice score, 2D average symmetric surface distance (ASSD), and Hausdorff robust (95 percentile) metrics. A network architecture similar to Dudovitch et al. \cite{dudovitch2020deep} was utilized with the Keras and Tensorflow frameworks.\\[01.ex]
\indent
{\bf Study 1}: We evaluate the performance of the ST method alone on the fetal body segmentation ID and OOD data. Baseline networks with 6 training examples were first trained with cases randomly selected from the 30 FIESTA  scans dataset. The ST was applied to the remaining 24 cases. Performance of the automatic ST method was tested on 68 ID FIESTA scans and 58 OOD TRUFI scans. Since sometimes longer training may result in improved performance \cite{fadida2022partial}, the performance of the ST method was also compared to that of fine-tuned networks without ST using learning rate restarts optimization scheme.

\begin{figure}[t]
\centering
\includegraphics[height=5cm, width=12.2cm]{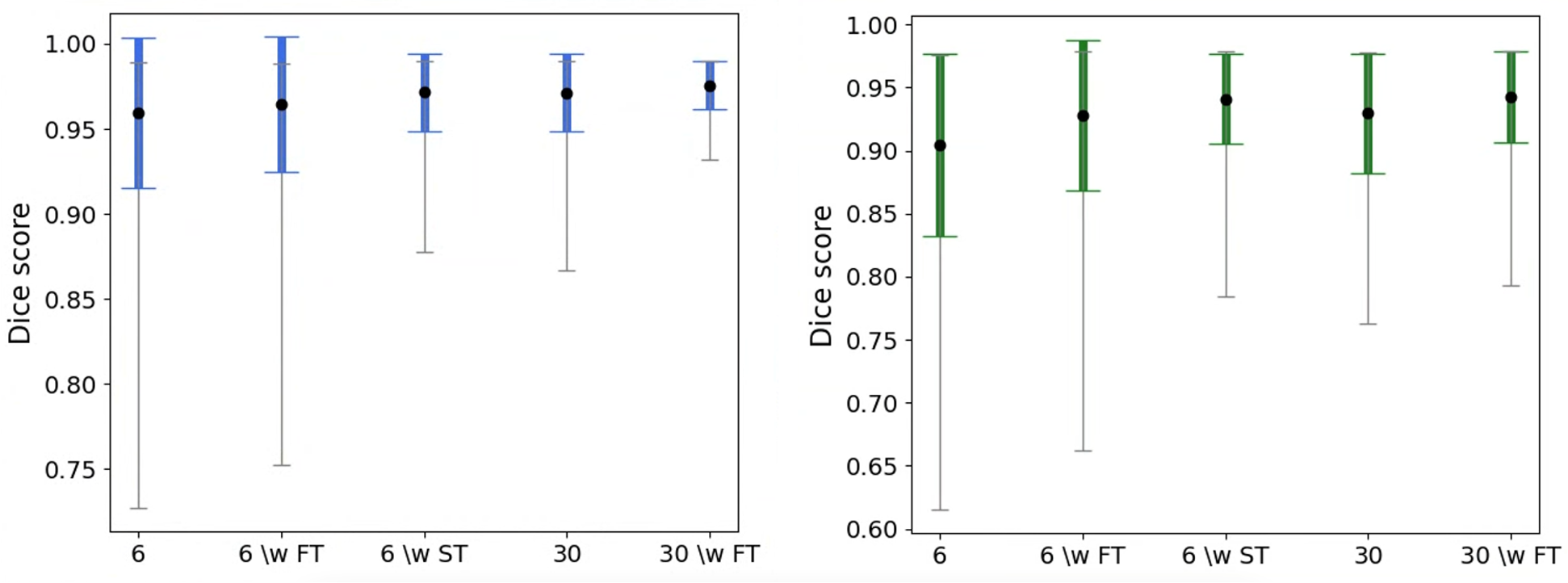}
\caption{Experimental results for the self training method on FIESTA ID and TRUFI OOD data averaged across four runs. Dots represent the mean Dice scores; grey bars show the minimum and the maximum; standard deviation: ID data (blue bar), OOD data (green bar). FT: Fine-tuning, ST: Self-training. Comparison between (x-axis): 1) 6 training examples (baseline); 2) 6 training examples with FT; 3) 6 training examples with ST 4) 30 training examples; 5) 30 training examples with FT}
\label{fig:sl_results_body}
\end{figure}

 For ID data, using self-training (ST) improved the Dice score from 0.964 to 0.971, the ASSD from 1.88 $mm$ to 1.52 $mm$ and the Hausdorff distance from 6.15 $mm$ to 4.8 $mm$. For the OOD data, the difference in performance was higher, with the self-training method improving the Dice score from 0.920 to 0.941, the ASSD from 4.7 $mm$ to 3.67 $mm$, and the Hausdorff distance from 21.01 $mm$ to 15.76 $mm$. Fig.~\ref{fig:sl_results_body} shows the results with the Dice score. The self-training also improved the standard deviation from 0.04 to 0.023 and from 0.06 to 0.036 for ID and OOD data respectively, and achieved the same performance of a network trained on manually segmented 30 cases for both ID and OOD data.

{\bf Study 2}: We evaluate the performance of the combined AL and ST method for FIESTA to TRUFI sequence transfer for fetal body segmentation. We subsequently performed sequence transfer for the TRUFI sequence using the ST network from the previous experiment as the initial network. To evaluate the differences in performance between the four runs, this time we calculated the average, STD, minimum, and maximum for the average performance of each one of the runs.
We added either: 1) 3 additional randomly selected TRUFI scans; 2) 3 additional AL TRUFI scans; 3)  3 additional AL TRUFI scans in combination with ST, or; 4) 2 additional AL TRUFI scans in combination with ST and with border slices annotations. We also obtained results for: 5) a network trained without the automatic ST step using 3 additional TRUFI AL scans and with ST that includes border slices annotations, and; 6) a baseline of 30 fully annotated TRUFI scans.

\begin{table}[t]
\centering
\caption{Fetal body segmentation results comparison with the Dice, Hausdorff 95 percentile and 2D ASSD metrics between: 1) 6 FIESTA cases and additional 3 randomly selected TRUFI scans; 2) 6 FIESTA cases and 3 additional actively selected TRUFI scans; 3)  6 FIESTA cases and 3 additional actively selected TRUFI scans in combination with ST; 4) 6 FIESTA cases and 2 additional actively selected TRUFI scans in combination with ST and with additional annotations of border slices. 5) a network trained without the automatic ST step using 6 FIESTA cases and 3 additional TRUFI actively selected scans with ST and border slices annotations, and; 6) a baseline of 30 fully annotated TRUFI scans. For (1-5), average $\pm$ STD and min-max range are listed for the four runs.}
\label{table:body_results}
\begin{tabular}{|l|c|c|c|}
\hline
\multicolumn{1}{|c|}{\textbf{}}                                                                                                                        & \multicolumn{1}{c|}{\textbf{Dice}} & \multicolumn{1}{c|}{\textbf{Hausdorff}} & \multicolumn{1}{c|}{\textbf{ASSD}} \\ \hline
\multirow{2}{*}{\textbf{1. 6 FIESTA + 3 TRUFI Random}}                                                                                                    & $0.957\pm0.005$                        & $8.36\pm0.35$                               & $2.51\pm0.28$                          \\ \cline{2-4} 
                                                                                                                                                       & 0.950-0.963                        & 7.76-8.67                               & 2.29-2.97                          \\ \hline
\multirow{2}{*}{\textbf{2. 6 FIESTA + 3 TRUFI AL}}                                                                                                        & $0.957\pm0.003$                         & $8.85\pm0.74$                               & $2.53\pm0.14$                          \\ \cline{2-4} 
                                                                                                                                                       & 0.953-0.961                        & 7.6-9.58                                & 2.34-2.74                          \\ \hline
\multirow{2}{*}{\textbf{3. 6 FIESTA + 3 TRUFI AL ST}}                                                                                                     & $0.959\pm0.002$                        & $7.84\pm1.16$                               & $2.45\pm0.15$                          \\ \cline{2-4} 
                                                                                                                                                       & 0.956-0.962                        & 6.29-9.54                               & 2.23-2.65                          \\ \hline
\multirow{2}{*}{\textbf{\begin{tabular}[c]{@{}l@{}}4. 6 FIESTA + 2 TRUFI AL ST \\ \textbackslash{}w borders\end{tabular}}}                                                                          & \textbf{$\textbf{0.961}\pm0.001$}               & \textbf{$\textbf{6.40}\pm0.41$}                      & \textbf{$\textbf{2.20}\pm0.14$}                 \\ \cline{2-4} 
                                                                                                                                                       & \textbf{0.960}-0.963                        & 5.73-\textbf{6.78}                               & 2.00-\textbf{2.38}                          \\ \hline
\multirow{2}{*}{\textbf{\begin{tabular}[c]{@{}l@{}}5. 6 FIESTA + 3 TRUFI AL ST \\ \textbackslash{}w borders \textbackslash{}wo ST iteration\end{tabular}}} & $0.959\pm0.003$                        & $8.09\pm1.14$                               & $2.37\pm0.25$                          \\ \cline{2-4} 
                                                                                                                                                       & 0.955-0.963                        & 6.23-9.08                               & 1.95-2.59                          \\ \hline
\rowcolor[HTML]{E9E7E7}
\textbf{6. 30 Baseline scans}                                                                                                                                   & 0.964                              & 6.99                                    & 2.14                               \\ \hline
\end{tabular}
\end{table}

Table~\ref{table:body_results} lists the results. The best results were achieved using only two additional labeled TRUFI scans with border slices annotations, yielding a mean Dice score of 0.961, close to that of the 30 fully annotated TRUFI scans baseline of 0.964. There was almost no difference between active versus random scan selection. Using the combined AL and ST method without the automatic ST step yielded very good results, although slightly worse than those of a network trained following the ST step, despite an additional annotated TRUFI scan. The worse network performance without the automatic ST iteration yielded an average Dice score of 0.955 vs. 0.960 using the ST iteration.

{\bf Study 3}: This study evaluates ST method and the combined AL and ST method on the high-variability data of placenta segmentation with both FIESTA and TRUFI sequences. Four initial networks were trained with 10 randomly selected training examples out of the 50 FIESTA and TRUFI scans, with the rest 40 cases used for ST and AL. The combined FIESTA and TRUFI data was split into 50/4/34 train/validation/test cases. An additional 24 scans were used for OOD data validation. Again, we evaluated the average, STD, minimum, and maximum for the average performances of the four runs. We compared between six scenarios: 1) a baseline network trained on 10 scans; 2) a network trained on 10 initial scans and additional ST scans; 3) a network trained on 10 initial scans and additional 5 randomly selected scans; 4) a network trained with 10 initial scans and additional 5 AL scans; 5) a network trained with 10 initial scans, 5 AL scans, and additional ST scans; 6) a network trained on 50 scans.

Table~\ref{table:placenta_results} lists the results. Using ST alone improved segmentation results from a mean Dice score of 0.806 to 0.819 for ID data and from 0.641 to 0.703 for OOD data. Using actively selected scans performed better than random selection, with a much smaller STD, reaching the performance of a network trained on 50 annotated scans. For ID data, the mean and minimum average Dice scores with actively selected scans were 0.827 and 0.819, respectively, and a mean and minimum average Dice scores of random selection were 0.821 and 0.809, respectively. For OOD data, the mean and minimum average Dice scores of actively selected scans were 0.710 and 0.661, compared to a mean of 0.688 and a minimum of 0.620 using random selection. 

Combining ST with AL did not improve the results, except for the OOD data Dice score STD, which was the lowest of all setups, and a minimum Dice score of 0.691, which was the highest. It slightly hurt performance on ID data, indicating that ST might induce excessive noise in the pseudo-labels. Adding automatic ST iteration may improve the pseudo-labels quality, but the effect may be negligible as the results are already on-par with a network trained on 50 cases.



\begin{table}[t]
\centering
\caption{Placenta segmentation results for ID and OOD data using the Dice, Hausdorff percentile 95, and 2D ASSD metrics.}
\label{table:placenta_results}
\begin{tabular}{|l|l|ccc|ccc|}
\hline
\multicolumn{1}{|c|}{}     & \multicolumn{1}{c|}{} & \multicolumn{3}{c|}{\textbf{ID Data}}                                                                                   & \multicolumn{3}{c|}{\textbf{OOD Data}}                                                                                   \\ \hline
\textbf{}                  & \textbf{}             & \multicolumn{1}{r|}{\textbf{Dice}}                 & \multicolumn{1}{r|}{\textbf{Hausdorff}}            & \textbf{ASSD} & \multicolumn{1}{r|}{\textbf{Dice}}                 & \multicolumn{1}{r|}{\textbf{Hausdorff}}            & \textbf{ASSD}  \\ \hline
\textbf{1. 10 Baseline}    & \textbf{Mean}         & \multicolumn{1}{r|}{0.806}                         & \multicolumn{1}{r|}{20.98}                         & 7.64          & \multicolumn{1}{r|}{0.641}                         & \multicolumn{1}{r|}{35.53}                         & 17.65          \\ \hline
\textbf{}                  & \textbf{STD}          & \multicolumn{1}{r|}{0.003}                         & \multicolumn{1}{r|}{1.13}                          & 0.26          & \multicolumn{1}{r|}{0.021}                         & \multicolumn{1}{r|}{8.12}                          & 1.62           \\ \hline
\textbf{}                  & \textbf{Min}          & \multicolumn{1}{r|}{0.803}                         & \multicolumn{1}{r|}{19.08}                         & 7.34          & \multicolumn{1}{r|}{0.628}                         & \multicolumn{1}{r|}{25.53}                         & 15.12          \\ \hline
\textbf{}                  & \textbf{Max}          & \multicolumn{1}{r|}{0.811}                         & \multicolumn{1}{r|}{22.00}                         & 8.03          & \multicolumn{1}{r|}{0.678}                         & \multicolumn{1}{r|}{46.46}                         & 19.48          \\ \hline
\textbf{2. 10 ST}          & \textbf{Mean}         & \multicolumn{1}{r|}{0.819}                         & \multicolumn{1}{r|}{19.75}                         & 7.11          & \multicolumn{1}{r|}{0.703}                         & \multicolumn{1}{r|}{29.23}                         & 18.86          \\ \hline
\textbf{}                  & \textbf{STD}          & \multicolumn{1}{r|}{0.010}                         & \multicolumn{1}{r|}{2.24}                          & 0.35          & \multicolumn{1}{r|}{0.028}                         & \multicolumn{1}{r|}{7.08}                          & 1.93           \\ \hline
\textbf{}                  & \textbf{Min}          & \multicolumn{1}{r|}{0.806}                         & \multicolumn{1}{r|}{16.95}                         & 6.65          & \multicolumn{1}{r|}{0.656}                         & \multicolumn{1}{r|}{23.22}                         & 16.49          \\ \hline
\textbf{}                  & \textbf{Max}          & \multicolumn{1}{r|}{0.833}                         & \multicolumn{1}{r|}{23.21}                         & 7.47          & \multicolumn{1}{r|}{0.727}                         & \multicolumn{1}{r|}{40.74}                         & 21.56          \\ \hline
\textbf{3.  10+5 Random} & \textbf{Mean}         & \multicolumn{1}{r|}{0.821}                         & \multicolumn{1}{r|}{19.30}                         & 7.02          & \multicolumn{1}{r|}{0.688}                         & \multicolumn{1}{r|}{29.29}                         & 13.68          \\ \hline
\textbf{}                  & \textbf{STD}          & \multicolumn{1}{r|}{0.008}                         & \multicolumn{1}{r|}{1.25}                          & 0.33          & \multicolumn{1}{r|}{0.039}                         & \multicolumn{1}{r|}{6.64}                          & 0.90           \\ \hline
                           & \textbf{Min}          & \multicolumn{1}{r|}{0.809}                         & \multicolumn{1}{r|}{17.81}                         & 6.68          & \multicolumn{1}{r|}{0.620}                         & \multicolumn{1}{r|}{23.81}                         & 12.50          \\ \hline
                           & \textbf{Max}          & \multicolumn{1}{r|}{0.832}                         & \multicolumn{1}{r|}{21.06}                         & 7.46          & \multicolumn{1}{r|}{0.717}                         & \multicolumn{1}{r|}{40.35}                         & 14.91          \\ \hline
\textbf{4. 10+5 AL}      & \textbf{Mean}         & \multicolumn{1}{r|}{\textbf{0.827}}                & \multicolumn{1}{r|}{\textbf{17.34}}                & \textbf{6.59} & \multicolumn{1}{r|}{\textbf{0.710}}                & \multicolumn{1}{r|}{\textbf{25.49}}                & 12.88          \\ \hline
\textbf{}                  & \textbf{STD}          & \multicolumn{1}{r|}{0.006}                         & \multicolumn{1}{r|}{1.10}                          & 0.46          & \multicolumn{1}{r|}{0.030}                         & \multicolumn{1}{r|}{\textbf{1.76}}                 & 1.35           \\ \hline
\textbf{}                  & \textbf{Min}          & \multicolumn{1}{r|}{0.819}                         & \multicolumn{1}{r|}{16.52}                         & 5.96          & \multicolumn{1}{r|}{0.661}                         & \multicolumn{1}{r|}{23.59}                         & 11.52          \\ \hline
\textbf{}                  & \textbf{Max}          & \multicolumn{1}{r|}{0.835}                         & \multicolumn{1}{r|}{19.21}                         & 7.11          & \multicolumn{1}{r|}{0.743}                         & \multicolumn{1}{r|}{27.98}                         & 15.12          \\ \hline
\textbf{5. 10+5 AL + ST}   & \textbf{Mean}         & \multicolumn{1}{r|}{0.822}                         & \multicolumn{1}{r|}{19.69}                         & 6.90          & \multicolumn{1}{r|}{\textbf{0.711}}                & \multicolumn{1}{r|}{26.50}                         & \textbf{12.59} \\ \hline
\textbf{}                  & \textbf{STD}          & \multicolumn{1}{r|}{0.006}                         & \multicolumn{1}{r|}{1.25}                          & 0.12          & \multicolumn{1}{r|}{\textbf{0.014}}                & \multicolumn{1}{r|}{3.35}                          & \textbf{0.92}  \\ \hline
\textbf{}                  & \textbf{Min}          & \multicolumn{1}{r|}{0.813}                         & \multicolumn{1}{r|}{18.33}                         & 6.76          & \multicolumn{1}{r|}{\textbf{0.691}}                & \multicolumn{1}{r|}{21.82}                         & 11.48          \\ \hline
                           & \textbf{Max}          & \multicolumn{1}{r|}{0.829}                         & \multicolumn{1}{r|}{21.68}                         & 7.09          & \multicolumn{1}{r|}{0.728}                         & \multicolumn{1}{r|}{31.29}                         & 13.82          \\ \hline
\rowcolor[HTML]{E6E2E2} 
\textbf{6.  50 Baseline}   & \textbf{Mean}         & \multicolumn{1}{r|}{\cellcolor[HTML]{E6E2E2}0.825} & \multicolumn{1}{r|}{\cellcolor[HTML]{E6E2E2}18.34} & 7.16          & \multicolumn{1}{r|}{\cellcolor[HTML]{E6E2E2}0.710} & \multicolumn{1}{r|}{\cellcolor[HTML]{E6E2E2}24.32} & 13.81          \\ \hline
\end{tabular}
\end{table}
\section{Conclusion} 
We presented a novel TTA-based framework for combined training with AL and ST in one iteration. The ST uses soft labels acquired from TTA results and border slices annotations when appropriate. Our experimental results show the effectiveness of the ST method alone for both low and high variability data regimes of fetal body and placenta segmentation, respectively. The combined iteration of AL and ST was effective when applied to the low data variability of the fetal body segmentation, especially with border slices annotations which have a low annotation cost, yielding a Dice score of 0.961 for sequence transfer using 6 examples from the original sequence and only 2 examples from the new sequence, close to that of 30 annotated cases from the new sequence. In this scenario of low variability data, AL did not improve upon random selection. However,  AL was effective for the high data variability setup, but adding ST to AL resulted in slight performance decline for ID data, indicating that ST might induce excessive noise in the pseudo-labels. In this case, AL may be better without ST. This suggests that for methods that combine AL with ST or CR SSL, a supervised network iteration with AL cases can be beneficial before the semi-supervised fine-tuning. Future work can explore this for different data variability settings.
%
%
%
%

\end{document}